\documentclass{article}

\usepackage{microtype}
\usepackage{graphicx}
\usepackage{subcaption}
\usepackage{booktabs}       

\usepackage{hyperref}

\usepackage[preprint]{icml2026}

\usepackage[utf8]{inputenc} 
\usepackage[T1]{fontenc}    
\usepackage{url}            
\usepackage{amsfonts}       
\usepackage{amsmath}        
\usepackage{amssymb}
\usepackage{mathtools}
\usepackage{amsthm}
\usepackage{bm}             
\usepackage{nicefrac}       
\usepackage{xcolor}         
\usepackage{algorithm}
\usepackage{algorithmic}
\usepackage{braket}
\usepackage{multirow}

\usepackage[capitalize,noabbrev]{cleveref}

\newcommand{\qfim}{\mbox{QFIM}}
\newcommand{\vqc}{\mbox{VQC}}
\newcommand{\mathsc}[1]{{\normalfont\textsc{#1}}}
\definecolor{errgray}{gray}{0.55}
\newcommand{\err}[1]{\textcolor{errgray}{$\,\pm\,$#1}}

\icmltitlerunning{\textsc{Quiver}: Quantum-Informed Views for Enhanced Representations in Large ML Models}

\begin{document}

\twocolumn[
  \icmltitle{\textsc{Quiver}: \textsc{Qu}antum-Informed Views for Enhanced Representations in Large Machine Learning Models}

  \icmlsetsymbol{equal}{*}

  \begin{icmlauthorlist}
    \icmlauthor{Aritra Bal}{itp,etp}
    \icmlauthor{Michael Binder}{etp}
    \icmlauthor{Markus Klute}{etp}
    \icmlauthor{Benedikt Maier}{imperial}
    \icmlauthor{Michael Spannowsky}{itp,iqmt}
  \end{icmlauthorlist}

  \icmlaffiliation{itp}{Institute of Theoretical Physics (ITP), KIT, Karlsruhe, DE}
  \icmlaffiliation{etp}{Institute of Experimental Particle Physics (ETP), Karlsruhe Institute of Technology (KIT), Karlsruhe, DE}
  \icmlaffiliation{imperial}{Blackett Laboratory, Imperial College of Science and Technology, London, UK}
  \icmlaffiliation{iqmt}{Institute for Quantum Materials and Technologies, KIT, Karlsruhe, DE}

  \icmlcorrespondingauthor{Aritra Bal}{aritra.bal@kit.edu}

  \icmlkeywords{Machine Learning, Quantum Computing, High Energy Physics, Variational Quantum Circuits, ICML}

  \vskip 0.3in
]

\printAffiliationsAndNotice{}  

\begin{abstract}
  Large machine learning models benefit substantially from multimodal inputs that provide a complementary view of the same example. We introduce \textsc{Quiver} (\textbf{QU}antum-\textbf{I}nformed \textbf{V}iews for \textbf{E}nhanced \textbf{R}epresentations), a paradigm that enriches classical data-driven features with a \emph{quantum Fisher view}: a geometrically motivated, basis-independent summary of higher-order correlations captured by a variational quantum circuit (VQC) trained to perform the same task. Unlike classical feature augmentation, the quantum Fisher information matrix encodes the intrinsic geometry of the learned quantum state manifold. While this feature map, motivated by quantum information theory, is ordinarily non-trivial to model classically, it can surface statistical structure that additional classical data or model capacity finds difficult to learn. This makes the quantum Fisher view a genuinely complementary modality rather than a redundant one. We demonstrate that  \textsc{Quiver} improves standard performance metrics on two benchmark datasets from very different fields: QM9 for predicting molecule properties, and \textsc{JetClass} for predicting jet flavor at the Large Hadron Collider (LHC). The core contribution, however, is domain-agnostic: the quantum Fisher view can be fused into a broad class of model architectures via targeted modifications to the base architecture, to incorporate information about the quantum geometry of the problem. These results demonstrate that quantum-geometric features, extracted from simulated variational circuits, can deliver measurable value for standard machine learning tasks, well before the advent of fault-tolerant quantum hardware.
\end{abstract}



\section{Introduction}
\label{sec:introduction}

Multi-million parameter models play an increasingly important role in scientific analysis of extremely high-dimensional data. Architectures such as graph neural networks (GNNs)~\citep{gilmer2017mpnn,battaglia2018relational,scarselli2009graph} and transformers~\citep{vaswani2017attention} have become standard tools in this regard. Two domains where this paradigm has driven significant methodological
progress are high-energy physics (HEP) and molecular chemistry. In both cases the inputs are high-dimensional structured objects, particles in a jet or atoms in a molecule, yet the models that process them are trained exclusively on classical representations of these systems. While these representations are highly effective, they are ultimately restricted to correlations that can be efficiently expressed within standard feature spaces and architectures.

Critical tasks in the context of major experiments such as the Large Hadron Collider (LHC), among others, are performed by such machine learning models, two prominent examples being \emph{jet flavor classification}, and \emph{anomaly detection} for new-physics searches. Jets are collimated sprays of hadrons produced when a high-energy quark or gluon fragments and hadronizes; identifying their parton-level origin (e.g.\ light quark vs.\ gluon, or boosted top quark vs.\ QCD background) is a central ingredient of essentially every analysis at the LHC. State-of-the-art jet taggers represent each jet as a \emph{point cloud} of its constituent particles, each carrying kinematic features~\citep{komiske2019efn,qu2020particlenet,qu2022particletransformer}, in addition to other information. Similar principles are used to design model-agnostic anomaly-detection pipelines aimed at uncovering physics beyond the Standard Model~\citep{nachman2020anode,kasieczka2021lhcolympics}.

While such architectures excel at exploiting kinematic correlations in the particle-level representation of a jet, they are limited by the structure of the feature space in which these correlations are expressed. In particular, higher-order and non-local correlations must be learned implicitly through model capacity rather than being directly exposed to the model. This motivates the exploration of alternative representations that can surface such structure more directly. Jets arise from a coherent branching process governed by quantum chromodynamics, and consequently exhibit rich correlation patterns among their constituents. In practice, however, these objects are represented for machine learning through classical feature constructions, usually kinematic, which can obscure or compress precisely those multi-particle correlations that are most discriminative—for example, between a color-singlet $W$ jet and a color-connected QCD jet. This motivates the exploration of alternative representations that make such structure more directly accessible. In this work, we operationalize this idea through the notion of \emph{quantum views} of a jet: embeddings of classical data into a Hilbert space that expose its geometric and correlation structure in ways not readily captured by standard kinematic summaries.

This structural limitation extends equally to molecular chemistry, where a central task is the prediction of quantum-mechanical molecular properties from structure, with targets such as the HOMO-LUMO gap, dipole moment, isotropic polarizability and atomization energy available in benchmark datasets like QM9~\citep{ramakrishnan2014qm9}. GNNs and transformers tailored to molecular graphs and atomic point clouds, SchNet~\citep{schutt2017schnet}, MPNN~\citep{gilmer2017mpnn}, DimeNet~\citep{klicpera2020dimenet} and equivariant successors, now define the state of the art on these benchmarks. These targets often depend on complex, highly correlated interactions that are only indirectly reflected in classical structural descriptors. As a result, models must infer these relationships from data rather than accessing a representation in which such correlations are naturally organized.

We address this limitation by introducing a \emph{quantum Fisher view} of the input in which we map classical data into a parameterized quantum state via a variational quantum circuit (VQC)~\citep{cerezo2021vqa}, and extract the associated quantum Fisher information matrix (QFIM)~\citep{liu2020qfimreview,meyer2021fisherqml,abbas2021powerqnn}. This construction does not require an assumption that the underlying system is quantum or that the encoding reflects a physical quantum state. Rather, it provides a principled mapping of classical data into a Hilbert space where geometric structure, in particular, sensitivity and higher-order correlations, can be probed through the induced metric.

On this basis, we propose \textsc{Quiver}, a paradigm that fuses the quantum Fisher view with the classical view of the same input. \textsc{Quiver} is deliberately architecture-agnostic: for transformer backbones we condition the model through cross-attention between the quantum and classical modalities, and for GNN backbones we modulate the learned graph messages by features derived from the \qfim{}. We show that \textsc{Quiver} delivers consistent improvements over classical baselines of similar or greater complexity, including the Particle Transformer~\citep{qu2022particletransformer}, a $>\!2$\,M-parameter state-of-the-art jet tagger used in LHC analyses, and DimeNet++, a state-of-the-art model on QM9 property prediction.

\section{Motivating \textsc{Quiver}: A Mathematical Background}
\label{sec:qfv}

In this section, we provide a short mathematical overview of a \vqc{} and the \qfim{}. Thereafter, we describe the quantum encoding used for the HEP task, and the novel quantum encoding we develop for the molecular chemistry application. All quantum circuit operations described in this paper were simulated classically using the quantum simulator library \textsc{PennyLane}~\citep{pennylane}.

\subsection{Variational quantum circuits}
\label{subsec:VQC}
A \vqc{} is a parameterized unitary $U(\bm{\theta})$ acting on a fixed reference state of $N$ qubits~\citep{cerezo2021vqa}. With the standard initialization $|0\rangle^{\otimes N}$, the circuit prepares first an input encoding:
\begin{equation}
    |\psi(\bm{\Theta})\rangle \;=\; U(\bm{\Theta})\,|0\rangle^{\otimes N}, \qquad \bm{\Theta} \in \mathbb{R}^{P},
    \label{eq:vqc-state}
\end{equation}
where the $P$ angles $\bm{\Theta} = (\Theta_1, \dots, \Theta_P)$ are functions of the inputs: $\bm{\Theta} = \bm{\Theta}(x)$, with $x$ a jet or a molecule. This encoding is typically followed by a series of entanglement operations and trainable single-qubit rotations $R(\theta)$, which together constitute a variational ansatz. The output of the circuit is obtained by measuring one or more qubit observables, yielding an expectation value that serves as the prediction of the \vqc{}.  

\subsection{The Quantum Fisher Information Matrix}
Now, because the map of inputs $\bm{\theta}\!\mapsto\!|\psi(\bm{\Theta,\bm\theta})\rangle$ is smooth, the resultant quantum states form a submanifold of pure states whose canonical Riemannian structure is the Fubini--Study metric~(\citep{provost1980riemannian}). On pure states this metric coincides, up to an overall factor of four, with the \qfim{}~(\citep{braunstein1994statistical}):
\begin{equation}
    F_{ij}(\bm{\theta}) \;=\; 4\,\mathrm{Re}\!\left[\,\langle \partial_i \psi \,|\, \partial_j \psi \rangle - \langle \partial_i \psi \,|\, \psi \rangle\langle \psi \,|\, \partial_j \psi \rangle\,\right], \qquad 
    \label{eq:qfim}
\end{equation}

where $\partial_i \equiv \frac{\partial}{\partial \theta_i}$, yielding the line element
\begin{equation}
    \mathrm{d}s^2 \;=\; \tfrac{1}{4}\sum_{i,j} F_{ij}(\bm{\theta})\,\mathrm{d}\theta_i\,\mathrm{d}\theta_j,
    \label{eq:fubini-study}
\end{equation}
which measures the statistical distinguishability of two infinitesimally separated parameter points $\bm{\theta}$ and $\bm{\theta}+\mathrm{d}\bm{\theta}$ through the states they prepare. In our setting, the input enters through the data-dependent state preparation that precedes the trainable rotations, so $F_{ij}(\bm{\theta}; x)$ is an input-conditioned object: evaluated at a fixed reference $\bm{\theta}_0$, it characterizes how the encoded state of $x$ shapes the local geometry of the trainable-parameter manifold. This computation is tractable on existing classical simulators, using standard implementations such as that in \textsc{Pennylane}.

The diagonal $F_{ii}$ records how strongly the prepared state responds to a perturbation of $\theta_i$ alone, and so acts as a per-feature ``dynamic'' importance score. The off-diagonal $F_{ij}$ couple distinct parameters and are non-zero precisely when the two corresponding directions in input space act \emph{coherently} on overlapping qubit subsystems; they vanish whenever the two parameters drive factorized, independent parts of the state. Under an encoding of the form described in the two previous sections, this gives a direct relational reading: large off-diagonal entries between two qubits flag collective behavior of the corresponding input elements; while a nearly diagonal $F$ signals effectively independent contributions. The result is a compact relational tensor whose entries are directly consumable by attention layers, or by message-passing networks, as will now be detailed in the following sections.

\subsection{The $\mathbf{1P1Q}$ particle embedding}
\label{subsec:1p1q}
For jets we adopt the one-particle--one-qubit ($\mathrm{1P1Q}$) encoding of~\citet{Bal2025}, in which each reconstructed constituent is mapped to a dedicated qubit using its kinematic features, followed by two-qubit entanglement, and standard Pauli rotation operations. We represent each jet as an ordered set of its ten highest $p_\mathrm{T}$ constituents, using $(p_\mathrm{T},\eta, \phi)$ as the set of kinematic input features. Here, $p_\mathrm{T}$ is the transverse momentum (magnitude of momentum in the plane perpendicular to the collider beam axis), $\eta = -\ln(\tan(\theta/2))$ is the pseudorapidity, and $\theta,\phi$ are the zenith and azimuthal angles. We use the standard coordinate references in collider physics where the $\mathrm{Z}$ axis is defined as being along the collider beam direction. We omit further details of the circuit and embedding, this being amply described in ~\cite{Bal2025}.

\subsection{The ${\mathbf{2A2Q}}$ molecular embedding}
\label{subsec:2a2q}
For the QM9 molecular dataset, we design and use a novel two-atom--two-qubit embedding which we call $\mathrm{2A2Q}$. The objective is to regress $\Delta \epsilon = \epsilon_\mathrm{HOMO}-\epsilon_\mathrm{LUMO}$, defined as the energy difference between the highest occupied (HOMO) and lowest unoccupied molecular orbitals (LUMO), on the QM9 dataset. We represent each molecule as a $10$-qubit system, with one qubit assigned to each heavy atom. The unused qubits are populated with randomly sampled hydrogen atoms, and all remaining explicit hydrogen information is otherwise discarded.

Starting from the initial state $\ket{0}^{\otimes N}$, we first learn a per-atom embedding by applying $R_Y(w_{\mathrm{atom}}^{j})\ket{0}$ on each qubit $j$, where $R_Y$ denotes the Pauli rotation about the $Y$-axis and $w_{\mathrm{atom}}^{j}$ is a trainable parameter associated with the atomic species occupying qubit $j$. A naive one-atom--one-qubit encoding of Cartesian coordinates would introduce a dependence on the choice of reference frame, which is undesirable. To mitigate this, we combine the encoding and entanglement stages into a single pairwise operation, defined by the angles
\begin{equation}
    \begin{aligned}
        \omega_{1}^{(ij)} &= e_{d_1}\cdot\left(1 - \frac{d_{ij}}{d_{\mathrm{CUTOFF}}}\right)\cdot\cos(\theta_{ij}), \\
        \omega_{2}^{(ij)} &= e_{\mathrm{bond}}^{(ij)}\cdot\pi, \\
        \omega_{3}^{(ij)} &= e_{d_2}\cdot\left(1 - \frac{d_{ij}}{d_{\mathrm{CUTOFF}}}\right)\cdot\cos(\phi_{ij}),
    \end{aligned}
\end{equation}
followed by the two-qubit unitary
\begin{align}
    \mathcal{U}_{ij} = \big(I_{YY}(\omega_{3}^{(ij)})\,I_{ZZ}(\omega_{2}^{(ij)})\,I_{XX}(\omega_{1}
    ^{(ij)})&\big) \nonumber \\\big(R_{Y}(w_{\mathrm{atom}}^{i})\otimes R_{Y}(w_{\mathrm{atom}}^{j})\big)\ket{00}&,
\end{align}
where $e_{d_1}$ and $e_{d_2}$ are learnable scaling parameters, $e_{\mathrm{bond}}^{(ij)}$ is a learnable bond-type entanglement parameter, $d_{\mathrm{CUTOFF}} = 1.7~\text{\AA}$ is fixed, and $I_{XX}$, $I_{YY}$, $I_{ZZ}$ denote the Ising-type two-qubit interactions $\exp(-i\,\omega\,\sigma_{X}\otimes\sigma_{X}/2)$, $\exp(-i\,\omega\,\sigma_{Y}\otimes\sigma_{Y}/2)$, and $\exp(-i\,\omega\,\sigma_{Z}\otimes\sigma_{Z}/2)$, respectively. The pairwise distance $d_{ij}$ is frame-invariant by construction, while the pairwise zenith and azimuthal angles $\theta_{ij}$ and $\phi_{ij}$ retain a residual frame dependence; we accept this trade-off, as the objective of the small variational quantum circuit is not to achieve state-of-the-art performance in isolation. The entanglement block $\mathcal{U}_{ij}$ is applied only for atom pairs $(i,j)$ satisfying $d_{ij} < d_{\mathrm{CUTOFF}}$ and connected by a chemical bond. The pairwise stage is followed by a per-qubit trainable rotation sequence $R_{Z}\cdot R_{Y}\cdot R_{Z}$ with independent parameters on each qubit. Together, the atom embedding, conditional pairwise entanglement, and single-qubit rotations constitute one layer of the circuit, and we stack $N = 2$ such layers in the final architecture.

The prediction for the gap $\Delta\epsilon$ is extracted from the \vqc{} via measurement of the observable:
\begin{equation}
    \mathcal{H} = \sum_{i=1}^{N} c_i\, Z_i,
\end{equation}
where the index $i$ runs over all $N = 10$ qubits of the system, $Z_i$ denotes the Pauli-$Z$ operator acting on the $i$-th qubit, and $\{c_i\}$ are trainable coefficients. Since the HOMO--LUMO gap is strictly positive, the raw expectation value $\langle\mathcal{H}\rangle \in [-\sum_i |c_i|,\, \sum_i |c_i|]$ is shifted by $\sum_i |c_i|$ so that the predicted value lies in $[0,\, 2\sum_i |c_i|]$. The circuit parameters are optimized by minimizing the Huber loss~(\cite{huber1964,pytorch_huberloss_211}) between the predicted gap and the target value in units of $\mathrm{meV}$, which provides robustness to outliers by interpolating between an $\ell_2$ behavior for small residuals and an $\ell_1$ behavior for large ones.

\section{Adding the Quantum Views}
\label{sec:adding-qv}
\subsection{Jet Flavor Classification}
\label{subsec:jet_tasks}
We evaluate \textsc{Quiver} on the \textsc{JetClass} dataset of \cite{jetclass}, 
focusing on the binary classification task of distinguishing hadronic top-quark jets
($t\!\to\!Wb\!\to\!q\bar qb$) against the QCD multijet background.
As our classical baseline, we adopt the Particle Transformer~\citep{qu2022particletransformer},
a state-of-the-art model with approximately  $2.14\mathrm{M}$ parameters~\footnote{We use the official release~\citep{weaver-core} of Particle Transformer on GitHub for the baseline}, and incorporate the \qfim{}
into its attention mechanism via implicit cross-attention through sequence concatenation, as described in Section~\ref{subsubsec:jet_qfi_injection}.

Table~\ref{tab:jet-features} summarizes the per-particle feature sets.
The kinematic baseline uses the per-particle kinematic-only features relative to the jet axis; the full-feature
baseline additionally includes calorimeter energy deposits, particle
identification flags, and track impact-parameters.

\begin{table}[htbp]
\centering
\small
\setlength{\tabcolsep}{4pt}
\caption{Per-particle input features for the jet tagging experiments.}
\label{tab:jet-features}
\begin{tabular}{p{1.8cm}|p{5.5cm}}
\toprule
Scope & Features \\
\midrule
Kinematics & $\log(p_{T,\text{rel}})$,\quad $\Delta\eta$,\quad $\Delta\phi$ \\
\multirow{4}{*}{Full}
  & \\[-2pt]
  & all kinematic features above, \textbf{plus}: \\[2pt]
  & $\log p_T$, $\log E$, $\log E_\text{rel}$,\quad $\Delta R$, charge \\
  & particle-ID flags (e$^{\pm}$, $\mu^{\pm}$, charged/neutral hadron, $\gamma$), impact-parameter features \\
  & ($d_0$,\; $d_z$ and their errors) \\
\bottomrule
\end{tabular}
\end{table}

All models also receive four-vectors $(p_x, p_y, p_z, E)$ for the
Lorentz-vector pair embedding and $(\Delta\eta, \Delta\phi)$ as
spatial point coordinates. 

\subsubsection{QFIM representation}
\label{subsubsec:jet_qfi_defn}
Given the compute requirements of simulating large multi-qubit states, we restrict our implementation of the $\mathrm{1P1Q}$ encoding to a maximum of $N=10$ qubits, allowing us to use only the ten highest-$p_\mathrm{T}$ constituents of a given jet, with all subsequent per-particle information being discarded. For $10$ particles encoded under the $\mathrm{1P1Q}$ scheme with three
local rotation-gate parameters per qubit, the \qfim{} is a
$30\!\times\!30$ real symmetric matrix, stored in the data pipeline as
90 channels over 10 particle slots. Each particle slot $i$ receives
all 90 \qfim{} channels as its feature vector, which is embedded by a
Particle-Transformer-style MLP into a token of dimension 128 and
appended to the classical particle-token sequence. The transformer
therefore receives 20 tokens in total: 10 classical particle tokens
followed by 10 \qfim{} tokens.

\subsubsection{The \textsc{Quiver} Paradigm: \qfim{} injection}
\label{subsubsec:jet_qfi_injection}
Let $\{x_i\}_{i=1}^{P}$ be the particle features, $\{v_i\}_{i=1}^{P}$ be the four-vectors used for pairwise embedding and $\mathbf{Q}\in\mathbb{R}^{90\times10}$ be the reshaped \qfim{} with the second dimension representing each constituent particle of the jet.
We incorporate the \qfim{} into the Particle Transformer by embedding
the per-particle \qfim{} channels independently using a
Particle-Transformer-style MLP and appending the resulting tokens to
the classical particle sequence:
\begin{equation}
    \text{transformer input}
    = \bigl[k_1, \dots, k_P,\; q_1, \dots, q_P\bigr],
    \label{eq:pconcat}
\end{equation}
where $k_i = \mathrm{MLP}_\text{tok}(x_i) \in \mathbb{R}^{128}$ are
the classical particle tokens and $q_i =
\mathrm{MLP}_\text{\qfim{}}(\mathbf{Q}[:,i]) \in \mathbb{R}^{128}$
are the embedded \qfim{} tokens, with $\mathbf{Q}[:,i]$ denoting all
$90$ \qfim{} channels associated with particle slot $i$. The Lorentz-vector pair bias
is computed for the original particle sequence and zero-padded to the
doubled sequence length. Algorithm~\ref{alg:jet-qfim} details the
complete forward pass.

\begin{algorithm}[t]
\caption{\textsc{Quiver}: QFIM token injection into ParT.}
\label{alg:jet-qfim}
\footnotesize
\begin{algorithmic}
\REQUIRE $\{x_i\}_{i=1}^{P}$, $\{v_i\}_{i=1}^{P}$, $\mathbf{Q}$
\FOR{$i = 1, \dots, P$}
    \STATE $k_i \leftarrow \mathrm{MLP}_\text{tok}(x_i)$
    \STATE $q_i \leftarrow \mathrm{MLP}_\text{qfim}(\mathbf{Q}[:,i])$
\ENDFOR
\STATE $b_{ij} \leftarrow \mathrm{PairEmbed}(v_i, v_j)$
\STATE Transformer Input $\;\leftarrow\;
       [k_1, \dots, k_P,\; q_1, \dots, q_P]$
\end{algorithmic}
\end{algorithm}

With the architectural modifications of Algorithm \ref{alg:jet-qfim}, the \textsc{Quiver}-augmented Particle Transformer has a parameter count of $2.29\mathrm{M}$, a modest increase of $7\%$ over the original.

\subsection{Molecular Property Regression}
\label{subsec:molecule}
The task, as before, remains the regression of $\Delta \epsilon$. The quantum encoding was previously described in Section~\ref{sec:qfv}. The \qfim{} is computed per molecule and stored as a $10\!\times\!10$
grid of $6\!\times\!6$ sub-blocks, corresponding to 2 circuit layers
times 3 single-qubit rotations per qubit, resulting in a $60\!\times\!60$
matrix whose off-diagonal block $Q_{ij}$ captures the coherent coupling
between the rotation-gate parameter groups of qubits $i$ and $j$, which by the conditions of the $\mathrm{2A2Q}$ encoding necessarily corresponds to that between atoms $i$ and $j$.

\subsubsection{The \textsc{Quiver} paradigm: Quantum-informed edge-state rescaling}
\label{subsubsec:molecule_sota}
Rather than introducing an independent QFIM processing branch, which
risks an improvement by generic parameter capacity rather than physically aligned
information, we constrain the \qfim{} to act as a modulating factor on top of the existing baseline edge-state vectors. 

We apply this mechanism to
DimeNet++~\citep{dimenetpp}\footnote{The baseline uses the DimeNet++ implementation in the package \textsc{PyTorch}
Geometric~\citep{Fey2019PyG,dimenetpp_pyg}.}, an improvement to the original DimeNet~\citep{klicpera2020dimenet}, which operates on directed-edge
embeddings $x_{ij}^{(l)}$ updated by interaction blocks rather than explicit
node messages. This rescaling is applied after the initial embedding block
and after each interaction block, so the steps described in Algorithm \ref{alg:dimenet-qfim}
are realizable in DimeNet++'s native edge state. The \qfim{}-modulated rescaling results in a parameter size of $1.891\mathrm{M}$, a negligible increase over the original's $1.886\mathrm{M}$ parameters. We refer to this model as $\mathcal{Q}$DimeNet++.

\begin{algorithm}[t]
\caption{\textsc{Quiver} in DimeNet++: QFIM-gated edge-state rescaling.}
\label{alg:dimenet-qfim}
\footnotesize
\begin{algorithmic}
\REQUIRE atomic numbers $\{z_i\}$, positions $\{\mathbf{r}_i\}$, \qfim{} sub-blocks $\{Q_{ij}\}$, learnable scalar $\alpha$
\STATE \textsc{Construct}: DimeNet++ radius graph and geometric basis functions: $\mathrm{RBF}_{ij}$ and $\mathrm{SBF}_{kji}$.
\STATE $x_{ij}^{(0)} \leftarrow \textsc{Embed}^{\mathrm{DimeNet++}}(z_i, z_j, \mathrm{RBF}_{ij})$
\STATE $\widetilde{x}_{ij}^{(0)} \leftarrow \textsc{Rescale}(x_{ij}^{(0)}, Q_{ij}, \alpha)$ \label{step:rescale}
\STATE $o \leftarrow \textsc{Output}^{(0)}_{\mathrm{DimeNet++}}(\widetilde{x}_{ij}^{(0)}, \mathrm{RBF}_{ij})$
\FOR{$l = 1, \ldots, L$}
    \STATE $x_{ij}^{(l)} \leftarrow \textsc{Interaction}^{(l)}_{\mathrm{DimeNet++}}(\widetilde{x}_{ij}^{(l-1)}, \mathrm{RBF}_{ij}, \mathrm{SBF}_{kji})$
    \STATE $\widetilde{x}_{ij}^{(l)} \leftarrow \textsc{Rescale}(x_{ij}^{(l)}, Q_{ij}, \alpha)$ \COMMENT{{QFIM gate}}
    \STATE $o \leftarrow o + \textsc{Output}^{(l)}_{\mathrm{DimeNet++}}(\widetilde{x}_{ij}^{(l)}, \mathrm{RBF}_{ij})$ \COMMENT{pooling}
\ENDFOR
\STATE $\hat{y}_G \leftarrow \sum_{i \in G} o_i$ \COMMENT{graph-level sum readout}
\end{algorithmic}
\end{algorithm}

The edge-state rescaling of Step \ref{step:rescale}, Algorithm \ref{alg:dimenet-qfim} is implemented as a residual
multiplicative gate on the baseline directed edge state $x_{ij}^{(l)}
$ 
\begin{equation}
\label{eq:quiver_message}
\widetilde{x}_{ij}^{(l)} = \mathsc{Rescale}(x_{ij}^{(l)}, Q_{ij}, \alpha)
=
\bigg(1+\alpha \cdot \Theta(Q_{ij}) \bigg) x_{ij}^{(l)}.
\end{equation}

Here, $\alpha$ is a global learnable scalar initialized to zero, ensuring that the two networks are identical in the beginning. The
function $\Theta(Q_{ij})$ is a per-edge bounded scalar learnt using a
convolutional neural network (CNN) applied to the $6\times6$ \qfim{}
sub-block, followed by a scaling multilayer perceptron (MLP)~\citep{rosenblatt1958perceptron} with a final $\tanh$ activation; ensuring
$\Theta(Q_{ij})\in[-1,1]$. The exact details of the CNN and training
setup are provided in the appendix.

The inputs and prediction targets are standardized
using the statistics from the train sample, with subsequent de-standardization of the output before reporting the results. Ten independent seed initializations are run for each variant to obtain the error bars.

\section{Results}
\label{sec:results}

\subsection{Jet Flavor Classification}
\label{sec:results-jet}

The evaluation criteria and classical benchmark was previously described in Section \ref{subsec:jet_tasks}. We train both the Particle Transformer (ParT) and its \textsc{Quiver}-augmented variant with five independent random seed initializations, and report our results in Table \ref{tab:ttbar_comparison}, using the AUC score and the QCD background rejection rate as our performance metrics. The latter quantity is a standard quantity used in HEP, defined as $1/\epsilon_B$. We evaluate this at a top-tagging efficiency of $\epsilon_S = 0.5$, (where $\epsilon_B = \mathrm{FPR}$, $\epsilon_S = \mathrm{TPR}$). Each run, comprising the aforementioned five initializations, is carried out twice, once for the kinematic-only set of features and then again for the full set of available inputs, both described in Table \ref{tab:jet-features}. 

\begin{table}[!h]
\centering
\footnotesize
\caption{%
    Comparison of ParT and \textsc{Quiver} on the top tagging task.
}
\label{tab:ttbar_comparison}
\resizebox{\columnwidth}{!}{%
\begin{tabular}{|c|l|l|l|}
\hline
{Features} & {Method} & {$N$} & {Results} \\
\hline
\multirow{12}{*}{Kin.}
    & \multirow{6}{*}{ParT}
        & \multirow{2}{*}{$0.1\,\mathrm{M}$}
            & \rule{0pt}{2.5ex}$\mathrm{AUC} = 0.97140$\err{$0.00038$} \\
    & & & $1/\epsilon_B = 107$\err{$2$} \\
\cline{3-4}
    & & \multirow{2}{*}{\rule{0pt}{2.5ex}$0.5\,\mathrm{M}$}
            & \rule{0pt}{2.5ex}$\mathrm{AUC} = 0.97629$\err{$0.00015$} \\
    & & & $1/\epsilon_B = 146$\err{$3$} \\
\cline{3-4}
    & & \multirow{2}{*}{\rule{0pt}{2.5ex}$5\,\mathrm{M}$}
            & \rule{0pt}{2.5ex}$\mathrm{AUC} = 0.97832$\err{$0.00004$} \\
    & & & $1/\epsilon_B = 176$\err{$1$} \\[3pt]
\cline{2-4}
    & \multirow{6}{*}{\textsc{Quiver}}
        & \multirow{2}{*}{\rule{0pt}{2.5ex}$0.1\,\mathrm{M}$}
            & \rule{0pt}{2.5ex}$\mathrm{AUC} = 0.97368$\err{$0.00013$} \\
    & & & $1/\epsilon_B = 130$\err{$1$} \\
\cline{3-4}
    & & \multirow{2}{*}{\rule{0pt}{2.5ex}$0.5\,\mathrm{M}$}
            & \rule{0pt}{2.5ex}$\mathrm{AUC} = 0.97848$\err{$0.00045$} \\
    & & & $1/\epsilon_B = 191$\err{$9$} \\
\cline{3-4}
    & & \multirow{2}{*}{\rule{0pt}{2.5ex}$5\,\mathrm{M}$}
            & \rule{0pt}{2.5ex}$\mathrm{AUC} = 0.98070$\err{$0.00003$} \\
    & & & $1/\epsilon_B = 240$\err{$1$} \\
\hline
\multirow{12}{*}{Full}
    & \multirow{6}{*}{ParT}
        & \multirow{2}{*}{$0.1\,\mathrm{M}$}
            & $\mathrm{AUC} = 0.98875$\err{$0.00008$} \\
    & & & $1/\epsilon_B = 570$\err{$13$} \\
\cline{3-4}
    & & \multirow{2}{*}{\rule{0pt}{2.5ex}$0.5\,\mathrm{M}$}
            & \rule{0pt}{2.5ex}$\mathrm{AUC} = 0.99080$\err{$0.00017$} \\
    & & & $1/\epsilon_B = 921$\err{$13$} \\
\cline{3-4}
    & & \multirow{2}{*}{\rule{0pt}{2.5ex}$5\,\mathrm{M}$}
            & \rule{0pt}{2.5ex}$\mathrm{AUC} = 0.99235$\err{$0.00003$} \\
    & & & $1/\epsilon_B = 1306$\err{$8$} \\[3pt]
\cline{2-4}
    & \multirow{6}{*}{\textsc{Quiver}}
        & \multirow{2}{*}{\rule{0pt}{2.5ex}$0.1\,\mathrm{M}$}
            & \rule{0pt}{2.5ex}$\mathrm{AUC} = 0.98893$\err{$0.00005$} \\
    & & & $1/\epsilon_B = 590$\err{$7$} \\
\cline{3-4}
    & & \multirow{2}{*}{\rule{0pt}{2.5ex}$0.5\,\mathrm{M}$}
            & \rule{0pt}{2.5ex}$\mathrm{AUC} = 0.99095$\err{$0.00003$} \\
    & & & $1/\epsilon_B = 951$\err{$17$} \\
\cline{3-4}
    & & \multirow{2}{*}{\rule{0pt}{2.5ex}$5\,\mathrm{M}$}
            & \rule{0pt}{2.5ex}$\mathrm{AUC} = 0.99244$\err{$0.00003$} \\
    & & & $1/\epsilon_B = 1362$\err{$28$} \\
\hline
\end{tabular}%
}
\end{table}

 These results demonstrate that the \textsc{Quiver} paradigm contributes to improving the performance of even large classical state-of-the-art baseline model such as the Particle Transformer in terms of community-standard performance metrics.

\subsection{Molecular Property Regression}
\label{sec:results-mol}
We evaluate \textsc{Quiver} on the benchmark described in Section \ref{subsubsec:molecule_sota}, with the results reported as
mean $\pm$ standard deviation over ten independent seed initializations. 

\begin{figure}[htbp]
    \centering
    \includegraphics[width=\linewidth]{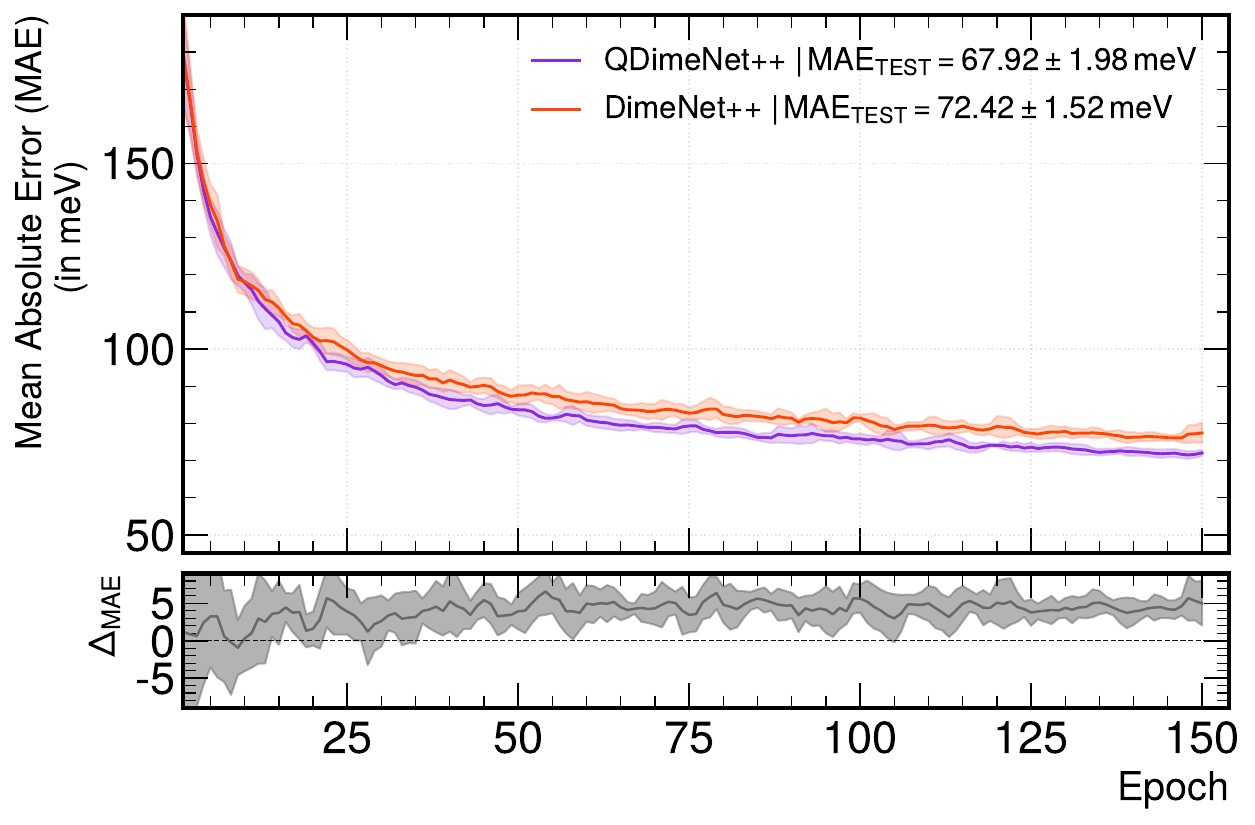}
    \caption{
    Validation MAE during training for the DimeNet++ baseline and the quantum-inspired $\mathcal{Q}$DimeNet++ model. 
    Solid lines show the mean across $N=10$ paired seeds (identical data splits and initialization RNG state across the two models; only the architecture differs), smoothed with a 3-epoch rolling window. 
    Shaded bands denote $\pm 1$ sample standard deviation across seeds (calculated using standard library functions with \texttt{ddof}=1). 
    The lower panel shows the per-epoch paired difference $\Delta\mathrm{MAE} = \mathrm{MAE}_{\text{DimeNet++}} - \mathrm{MAE}_{\mathcal{Q}\text{DimeNet++}}$ with the corresponding $\pm 1\sigma$ band over the same paired seeds; positive values indicate lower MAE for $\mathcal{Q}$DimeNet++.
}
    \label{fig:dimenet_qfim_difference}
\end{figure}

Figure~\ref{fig:dimenet_qfim_difference} shows the validation MAE
curves over training for both DimeNet++ variants.

The upper panel shows that $\mathcal{Q}$DimeNet++ consistently achieves
lower validation MAE than the baseline, with the gap opening early in
training and persisting through convergence. The lower panel shows the
per-epoch difference between DimeNet++ and $\mathcal{Q}$DimeNet++, which remains
positive throughout training.

Our results show that $\mathcal{Q}$DimeNet++ achieves a mean test MAE of $67.92 \pm 1.98\,\mathrm{meV}$
against $72.42 \pm 1.52\,\mathrm{meV}$ for the classical baseline, a relative reduction of $6.21\%$ achieved with a negligible parameter overhead
of $0.27\%$.

The mean paired difference is $\Delta\mathrm{MAE} = 4.50 \pm 2.46\,\mathrm{meV}$. A paired $t$-test across the ten seeds yields $t_9 = 5.78$ ($p < 10^{-3}$), confirming that the observed reduction is statistically significant and is not derived from seed-level noise. This serves as a demonstration
that the \qfim{} contributes genuine information rather than acting as a source of generic additional model capacity.

\section{Conclusion}

Our work introduced \textsc{Quiver}, an architecture-agnostic paradigm in which the quantum Fisher information matrix of a variational quantum circuit, evaluated on classical inputs, furnishes a geometry-aware view that complements the standard classical representation of the same example. The construction is deliberately decoupled from any assumption that the underlying system is quantum: it treats the embedding solely as a mapping into a Hilbert space whose induced Fubini--Study metric (equivalent to the \qfim{} up to an overall factor of four) exposes higher-order correlations that are not naturally organized within standard kinematic or structural feature spaces. Fusing this quantum Fisher view with the classical view, via targeted architectural modifications, yields consistent improvements over two state-of-the-art baselines on tasks drawn from very different domains. On the \textsc{JetClass} top-quark tagging benchmark, the QUIVER-augmented Particle Transformer improves both the AUC and the QCD background rejection $1/\epsilon_B$ at $\epsilon_S = 0.5$ across all training sample sizes and feature sets considered, at a $7\%$ parameter overhead. On the QM9 HOMO--LUMO gap regression task, $\mathcal{Q}$DimeNet++ reduces the test mean absolute error from $72.42 \pm 1.52$\,meV to $67.92 \pm 1.98$\,meV, a $6.21\%$ relative improvement obtained at a $0.27\%$ parameter overhead, with a mean paired difference $\Delta\mathrm{MAE} = 4.50 \pm 2.46$\,meV that remains positive within $\pm 1\sigma$ across the ten paired seeds. The persistence of these gains under negligible parameter overhead, and across distinct architectures, input modalities and physical symmetries, supports the interpretation that the QFIM supplies genuinely complementary information rather than acting as a source of generic model capacity. Taken together, these results indicate that quantum-geometric features extracted from classically simulated variational quantum circuits can deliver measurable value to large classical models today, decoupling the practical utility of quantum-informed representations from progress toward fault-tolerant hardware. 
\section{Limitations}
We provide a short discussion of the main limitations of this work, with a clear pathway for tackling these in future research. 

First, the computational overhead of simulating extremely large quantum circuits constrains us to use at most $10$ qubits for the \vqc{} of the $\mathrm{1P1Q}$ encoding, which contains the kinematic information of up to ten jet constituents, a limitation propagated therefore to the Particle Transformer benchmark. Even though most of the critical information required for jet flavor classification is often contained in these high-momentum constituents, this still results in a loss of performance, as compared to what would have been attained by using all $150$ jet constituent particles available in the \textsc{JetClass} dataset.

This limitation also affects the task of molecular property prediction: the absolute performance of our benchmark DimeNet++, and \textsc{Quiver}-augmented $\mathcal{Q}$DimeNet++ is marginally below that of the numbers reported in the original DimeNet++ paper~\citep{dimenetpp}, a necessary consequence of our setup operating on a restricted subset of up to $10$ atoms (these being mapped to $10$ qubits), resulting in the information contained in the remaining (hydrogen) atoms being lost. The results are interpretable as a clear methodological gain under identical conditions for both large-parameter models. The goal therefore remains to scale up these systems to more qubits by the usage of, for example, \textbf{HPC resources with multi-GPU nodes} for large-qubit quantum system simulations. 

Finally, a hybrid quantum-classical pipeline that simultaneously minimizes the parameters of both the precursor \vqc{} and the subsequent large neural model could in principle, converge to a global minimum with performance better than what is observed in the current iteration of this work. This remains among our goals, with the main technical challenge lying in optimizing the quantum circuit based on a measurement of its \qfim{} rather than an observable. 

\bibliographystyle{icml2026}
\bibliography{references}


\newpage
\appendix
\onecolumn

\section{Appendix: Jet Flavor Classification}
\subsection{Training setup and times}
 As mentioned in Table \ref{tab:ttbar_comparison}, we train the particle transformer per epoch on training sizes of $0.1\mathrm{M}$, $0.5\mathrm{M}$ and $5\mathrm{M}$ jets, equally divided between the two classes. 
We use a validation set of $1\mathrm{M}$ jets (equally balanced) and a held-out test set of $3.9\mathrm{M}$ jets. 
Training employs the Ranger optimizer (combining RAdam with LookAhead) with learning rate $\eta = 1 \times 10^{-3}$, batch size $512$, and a flat+decay learning rate schedule that maintains the initial learning rate for the first 70\% of training before exponentially decaying to $0.01\eta$ over the final 30\%. 
The model is trained with cross-entropy loss and early stopping is disabled, allowing models to train for the full $60$ epochs. 
For the largest training size of $5\mathrm{M}$ examples, the baseline Particle Transformer requires approximately $0.46\,\mathrm{h}$ per epoch, with its \textsc{Quiver}-augmented variant requiring $1.26\,\mathrm{h}$ for the same.

\section{Appendix: Molecular Property Regression}
\subsection{Training setup and times}
The QM9 dataset is partitioned into training, validation, and test subsets containing $65{,}390$, $13{,}078$, and $52{,}313$ molecules respectively, summing to the full corpus of $130{,}781$ samples. The \vqc{} is trained on a subset of $5{,}000$ molecules drawn from the training partition, with $1{,}000$ molecules taken from the validation partition for monitoring convergence. The classical DimeNet++ benchmark and the proposed $\mathcal{Q}\mathrm{DimeNet}{++}$ architecture are both trained on the remaining $60{,}390$ molecules and validated on the remaining $12{,}078$ molecules, with final performance reported on the held-out test set of $52{,}313$ molecules. This protocol guarantees that the classical and quantum-enhanced models are evaluated on identical test data, while ensuring that no sample seen by the \vqc{} during its pre-training stage is reused for validation or testing of the downstream graph network.

We train all DimeNet++ models with the \textsc{Adam} optimizer using an
initial learning rate of $10^{-3}$, batch size $128$, and zero weight
decay. No learning-rate schedule or decay is used. Models are trained
for at most $300$ epochs with early stopping on validation MAE: training
terminates when validation MAE fails to improve by at least
$0.25\,\mathrm{meV}$ for $30$ consecutive epochs. Since the HOMO--LUMO
gap target is standardized during training, this corresponds to a
normalized threshold of $1.95\times 10^{-4}$.

The training objective is the L1 loss, equivalent to MAE, applied to the
standardized HOMO--LUMO gap target. Validation and test MAEs are reported
after converting back to physical units using the training-set target
standard deviation.

Across the 10 seeds, DimeNet++ trained for
$196.9 \pm 30.9$ epochs on average, while $\mathcal{Q}$DimeNet++ trained
for $216.5 \pm 54.6$ epochs. Because runs were executed under parallel
GPU scheduling, wall-clock timings should be interpreted as approximate
runtime bounds rather than isolated architecture benchmarks. In this
setup, epochs completed within approximately $80\,\mathrm{s}$ for
DimeNet++ and $100\,\mathrm{s}$ for $\mathcal{Q}$DimeNet++.

For both cases, training was carried out on a single \textsc{Nvidia} $\mathrm{L40S}$ GPU with $48\,\mathrm{GB}$ of VRAM, on a locally available university compute cluster with 128 CPU cores.

\subsection{Technical Details}
The edge-state rescaling in Equation~\ref{eq:quiver_message}, proposed under the \textsc{Quiver} paradigm, operates on the $6\times 6$ \qfim{} sub-matrix that encodes the pairwise interaction between qubits $i$ and $j$. This sub-matrix is first processed by a two-dimensional convolutional layer with $16$ output channels and a kernel of size $3$, followed by a ReLU non-linearity. A global average pooling operation then collapses the spatial dimensions, and the resulting representation is flattened from $16$ channels into an $8$-dimensional vector, which is subsequently standardized by a LayerNorm operation to stabilize the learned embedding distribution. The normalized embedding is then passed through a scaling multilayer perceptron $s_{ij}$ consisting of a linear projection from $d_{\mathrm{QFIM}}$ to $\max(4, d_{\mathrm{QFIM}})$ hidden units, a SiLU activation, a second linear projection to a single scalar, and a final $\tanh$ non-linearity that bounds the output to $(-1, 1)$. The resulting scalar acts as a learned, edge-specific multiplicative gate on the message exchanged between qubits $i$ and $j$, allowing the model to attenuate or amplify each \qfim{}-derived interaction in a fully data-driven manner.

\end{document}